\let\JPSJKernelEndTabular\endtabular
\documentclass[fp]{jpsj3}
\renewcommand{\endtabular}{\JPSJKernelEndTabular\end{center}}

\usepackage[T1]{fontenc}
\usepackage{txfonts}
\usepackage{booktabs}
\usepackage{xurl}
\usepackage{xcolor}
\usepackage{multirow}

\usepackage[section]{placeins}

\newcommand{\myrev}[1]{#1}

\title{QUBO-based feature selection for MRI undersampling pattern design}

\author{Xiaoxin Huang$^1$\thanks{huang.xiaoxin.p6@dc.tohoku.ac.jp} and Masayuki Ohzeki$^{1,2,3,4}$}

\inst{$^1$Graduate School of Information Sciences, Tohoku University, Sendai 980-8579, Japan\\
$^2$Department of Physics, Institute of Science Tokyo, Tokyo 152-8551, Japan\\
$^3$Research and Education Institute for Semiconductors and Informatics, Kumamoto University, Kumamoto 860-0862, Japan\\
$^4$Sigma-i Co., Ltd., Tokyo 108-0075, Japan}

\abst{
We formulate undersampling mask design for Cartesian magnetic resonance imaging (MRI) as a feature subset selection problem.
Each sampling line is represented by a binary feature indicating whether it is included in each candidate mask.
Using candidate masks and their training-set evaluations, we estimate mutual information (MI) and conditional mutual information (CMI).
These estimates and a sampling-count penalty define a quadratic unconstrained binary optimization (QUBO) model.
The model is solved using simulated annealing and quantum annealing to determine a feature subset and its corresponding sampling mask.
We assess the selected subsets on the test set through zero-filled reconstruction in Fourier undersampling simulations using fastMRI knee magnitude reference images, with patient-disjoint training and test sets.
At all six tested target sampling rates from 15\% to 40\%, the resulting reconstructions have lower mean root mean square error than those obtained with four conventional sampling methods.
They also have higher mean structural similarity than those obtained using univariate mutual-information ranking at all tested rates.
These results support information-theoretic feature selection as an approach to MRI undersampling mask design under the tested zero-filled simulation settings.
}

\begin{document}
\maketitle

\section{Introduction}
Magnetic resonance imaging (MRI) obtains image information by acquiring spatially encoded signals.
In the conventional Cartesian Fourier imaging model, $k$-space data represent the Fourier coefficients of an image, which can be reconstructed from fully sampled data by an inverse Fourier transform.\nobreak\cite{gozcu2018}
However, full acquisition often requires long scan times, which can be challenging for people who experience claustrophobia or have difficulty remaining still for extended periods.
Reducing the amount of acquired $k$-space data through undersampling has therefore become an important approach to accelerating MRI.
Although undersampling shortens acquisition time, it also reduces the information available for image reconstruction, and a direct inverse Fourier transform generally produces aliasing artifacts.
Compressed sensing (CS) exploits the sparsity of MR images in suitable transform domains and combines undersampling with nonlinear reconstruction, providing a theoretical framework for recovering images from incomplete $k$-space data.\nobreak\cite{lustig2007}
Under undersampling, reconstruction performance depends both on the reconstruction method and on which $k$-space data are acquired.
For a given Cartesian $k$-space grid, the selected sampling positions can be represented by a binary mask whose elements indicate whether the corresponding positions are acquired.\nobreak\cite{gozcu2018}
Sampling mask design therefore requires selecting the $k$-space positions to retain under a given sampling budget while preserving reconstruction quality as far as possible.

A classical approach in compressed sensing MRI generates undersampling masks according to predefined sampling rules.
For example, Lustig et al. used pseudorandom variable-density sampling, in which sampling probabilities vary across $k$-space to increase both undersampling incoherence and sampling density in low-frequency regions.\nobreak\cite{lustig2007}
Such rules are specified in advance and generally neither use training data from a particular anatomy nor directly optimize sampling positions for a given reconstruction method.
To adapt sampling patterns to the imaged anatomy and reconstruction task, subsequent studies treated the sampling mask itself as an optimization variable.
These approaches use reconstruction performance on training images to determine which $k$-space positions to acquire under a given sampling budget.
G\"ozc\"u et al. searched for sampling masks based on average performance over training images for a specified reconstruction method.\nobreak\cite{gozcu2018}
LOUPE (Learning-based Optimization of the Under-sampling PattErn) jointly learns a sampling probability distribution and a reconstruction network, then generates masks from the learned distribution.\nobreak\cite{bahadir2020}
Weber et al. framed the problem as directly optimizing the mask at a given sampling rate, rather than improving reconstruction under a predefined undersampling mask.\nobreak\cite{weber2024}
In this formulation, reconstruction quality or downstream task performance is used to evaluate the acquired data.
These studies identify the central question in sampling design as determining which $k$-space data to retain under a limited sampling budget.
Here, we consider undersampling masks composed of Cartesian sampling lines with a fixed orientation, such that all $k$-space data on a line are retained or discarded together.
Designing a mask under a given sampling budget is therefore equivalent to selecting a subset of all available sampling lines.
We use this correspondence to investigate MRI sampling mask design from the perspective of feature selection (FS).
The selection state of each sampling line is treated as a feature.
The retained lines are determined by evaluating the statistical relationships between individual lines or their combinations and the reconstruction performance of candidate masks.

Feature selection methods are commonly classified as filter, wrapper, or embedded methods according to the relationship between the selection process and the downstream model.
Filter methods evaluate features using statistical or structural properties of the data, whereas wrapper methods evaluate candidate subsets through downstream model performance.
Embedded methods incorporate feature selection into model training.\nobreak\cite{chandrashekar2014}
Filter methods can construct selection scores by measuring statistical associations between features and a target variable; mutual information (MI) is a representative measure for this purpose.\nobreak\cite{peng2005}
MI captures statistical dependence between variables without requiring the relationship to be linear.
Univariate MI evaluates the association between each feature and the target separately.
Conditional mutual information (CMI) assesses this dependence conditional on other variables, providing a statistical basis for selection criteria that incorporate pairwise conditional relationships.\nobreak\cite{brown2012}

Once a selection criterion has been established, the features to be selected jointly must still be determined.
There are $\binom{n}{k}$ possible ways to select $k$ features from $n$ candidates, making exhaustive evaluation increasingly difficult as the problem grows.
When a selection criterion includes both univariate and pairwise scores, univariate ranking alone generally cannot identify its optimal subset.
Binary variables indicating whether features are selected allow univariate scores, pairwise scores, and a penalty for deviations from the target budget to be combined.
The resulting quadratic unconstrained binary optimization (QUBO) model expresses feature subset selection as a search for low-energy binary configurations.\nobreak\cite{mucke2023,nau2025}

QUBO provides a unified representation of the selection criterion and the cardinality constraint, together with a standard formulation for solving the combinatorial optimization problem by annealing.
A transformation between binary and spin variables gives an equivalent Ising energy function, whose low-energy spin configurations correspond to low-energy binary configurations of the QUBO model.\nobreak\cite{lucas2014}
Such energy minimization problems can be addressed using classical methods such as simulated annealing (SA).\nobreak\cite{kirkpatrick1983}
SA searches for low-energy solutions by gradually reducing an effective temperature while accepting higher-energy configurations with a specified probability.
QUBO can also be mapped to quantum annealing (QA) systems.
QA gradually reduces quantum fluctuations to drive the system toward low-energy states of the encoded target Ising model, offering another approach to combinatorial optimization.\nobreak\cite{kadowaki1998}

The connection between QUBO formulations and annealing has been applied to several types of feature selection problems.
M\"ucke et al. constructed QUBO models using feature importance and redundancy.\nobreak\cite{mucke2023}
Pranji\'{c} et al. used MI and CMI to evaluate features and their combinations, then assessed the selected features through excavator price prediction.\nobreak\cite{pranjic2026}
In medical imaging, Felefly et al. applied a QUBO model incorporating MI and CMI to feature selection for MRI radiomics.\nobreak\cite{felefly2023}
Nau et al. used a similar approach to select pixel subsets in medical images and evaluated the selections through image reconstruction.\nobreak\cite{nau2025}
These studies show that information-theoretic selection criteria can be organized into combinatorial optimization problems using QUBO, with the selected subsets evaluated through specific tasks.
Here, we examine whether this approach can be applied to MRI sampling line selection and assess the resulting masks through image reconstruction.

We investigate the design of MRI undersampling masks composed of Cartesian sampling lines as a feature selection problem under a given sampling budget.
Candidate masks serve as statistical samples, with the selection state of each sampling line represented as a binary feature.
A discrete target variable is constructed from the zero-filled reconstruction errors of candidate masks on the training set.
MI measures the statistical association between the selection state of an individual sampling line and the target variable, whereas bidirectional CMI characterizes conditional associations involving pairs of lines.
These information scores are combined with a sampling cardinality constraint to form a QUBO model.
We solve this model using SA and QA to obtain sampling line combinations and their corresponding masks.
The masks are then applied to an independent test set and compared with classical sampling methods and a univariate MI ranking baseline through image reconstruction.

\section{Methods}

\subsection{MRI undersampling and sampling-mask representation}

MRI $k$-space data and image data are related by the Fourier transform. In a $k$-space representation centered at zero frequency, the central region corresponds to low spatial frequencies and primarily reflects large-scale image structure and contrast, whereas the peripheral region corresponds to high spatial frequencies and contains more information about edges and fine details.\nobreak\cite{moratal2008} A sampling mask specifies the data locations retained during undersampling.
In the Fourier undersampling simulations considered here, let $u$ denote the reference MRI magnitude image, $\mathcal F$ the two-dimensional Fourier transform operator, $z$ the sampling-line selection vector, and $\mathcal M(z)$ the corresponding binary sampling mask. The undersampled observation is expressed as
\begin{equation}
b_z=A_z u=\mathcal M(z)\odot\mathcal F u,
\label{eq:new-observation}
\end{equation}
where $\mathcal F u$ is the $k$-space data corresponding to the reference image, $\odot$ denotes elementwise multiplication, and $A_z$ is the Fourier sampling operator defined by the mask. Mask values of 1 and 0 indicate retained and discarded data, respectively. The observation $b_z$ is represented on the full grid, with unsampled locations set to zero. Image reconstruction estimates $u$ from the retained data and their sampling locations.\nobreak\cite{gozcu2018}
We use Cartesian sampling lines in a fixed direction as the selection units, retaining or discarding all data on each line together. The binary variable $z_i$ indicates whether sampling line $i$ is selected: $z_i=1$ denotes retention and $z_i=0$ denotes exclusion. The selected set of sampling lines therefore determines the sampling mask.
For $N$ available sampling lines and a target sampling rate $r$, the target number of sampling lines is
\begin{equation}
k=\operatorname{round}(rN).
\label{eq:target-sampling-budget}
\end{equation}
Sampling-mask design can thus be formulated as a sampling-line subset-selection problem with a target cardinality of $k$.

\subsection{Feature representation and target-variable construction}

Estimating mutual information (MI) and conditional mutual information (CMI) requires joint samples of the feature and target variables. To obtain these statistical samples, we generate multiple candidate masks with the same sampling budget at each target sampling rate. Each candidate mask constitutes one statistical sample: the selection states of individual sampling lines form the binary features, and the discretized training-set reconstruction error of the candidate mask provides the target variable. These samples are subsequently used to estimate the MI between the selection state of each sampling line and the target variable, as well as the CMI conditioned on the selection state of another line.
At each target sampling rate, we generate $M$ candidate selection vectors
\(\{m^{(t)}\}_{t=1}^{M}\), where
\(m^{(t)}\in\{0,1\}^{N}\), with corresponding two-dimensional sampling masks
\(\mathcal M(m^{(t)})\). Every candidate satisfies the same target number of sampling lines:
\begin{equation}
\sum_{i=1}^{N}m_i^{(t)}=k.
\label{eq:candidate-budget}
\end{equation}

The candidate masks are generated using two rules: uniform random sampling over all available lines and low-frequency Gaussian sampling. Uniform random candidates select lines uniformly without replacement from the full set, without prescribing a contiguous low-frequency region.
The Gaussian candidates follow the construction of one-dimensional Gaussian variable-density sampling: contiguous lines around zero frequency are retained first, and the remaining lines are drawn from a Gaussian distribution centered at zero frequency until the target number of sampling lines is reached.\nobreak\cite{yiasemis2024}
The size of the contiguous low-frequency region scales with the target sampling budget. Together, these two candidate types provide sampling-line combinations with different frequency distributions for estimating the statistical association between line-selection states and training reconstruction performance.
For candidate $t$, we define
\begin{equation}
X_i^{(t)}=m_i^{(t)},
\label{eq:candidate-selection-state}
\end{equation}
where \(X_i^{(t)}\) is the binary selection state of sampling line $i$ in candidate mask $t$.
All candidate masks are evaluated using the same zero-filled reconstruction procedure. For reference image \(u_{p\ell}\), corresponding to slice $\ell$ of training patient $p$, the reconstruction obtained with candidate $t$ is
\begin{equation}
\widetilde u_{p\ell}^{(t)}
=\mathcal N\!\left(
\left|\mathcal F^{-1}\!\left[
\mathcal M(m^{(t)})\odot\mathcal F u_{p\ell}
\right]\right|\right),
\label{eq:candidate-zf}
\end{equation}
where \(|\cdot|\) denotes the pixelwise magnitude and \(\mathcal N\) denotes magnitude-image normalization. The difference between the reconstructed and reference images is measured by the root mean square error (RMSE), defined in Section~\ref{sec:reconstruction-evaluation}. Slice errors are first averaged within each patient and then averaged with equal weights across training patients, yielding the training reconstruction error of the candidate mask:
\begin{equation}
R^{(t)}
=\frac{1}{P_{\mathrm{tr}}}\sum_{p=1}^{P_{\mathrm{tr}}}
\frac{1}{L_p}\sum_{\ell=1}^{L_p}
\operatorname{RMSE}\!\left(u_{p\ell},\widetilde u_{p\ell}^{(t)}\right),
\label{eq:new-response}
\end{equation}
where \(P_{\mathrm{tr}}\) is the number of training patients and \(L_p\) is the number of training slices for patient $p$. This aggregation gives each patient equal weight in the evaluation of a candidate mask.
To estimate the information measures from discrete empirical frequencies, we discretize the training reconstruction error \(R^{(t)}\) by quantile binning to construct the target variable for feature selection. Following the quantile discretization of continuous targets used by Romero et al.,\nobreak\cite{romero2025} bin boundaries are determined from the empirical quantiles of the candidate masks' training errors. We define
\begin{equation}
Y^{(t)}=q_B\!\left(R^{(t)}\right),
\label{eq:response-discretization}
\end{equation}
where \(q_B\) is a quantile-binning map with a target of $B$ bins. The value \(Y^{(t)}\) identifies the quantile interval containing the training reconstruction error of candidate mask $t$.
The resulting statistical dataset is
\begin{equation}
\mathcal D_{\mathrm{mask}}
=\left\{\left(X_1^{(t)},\ldots,X_N^{(t)},Y^{(t)}\right)\right\}_{t=1}^{M}.
\label{eq:new-statistical-data}
\end{equation}
Each statistical sample corresponds to a candidate mask, with $X_i$ representing the selection state of a sampling line and $Y$ providing target information about training reconstruction performance. The estimated associations depend on the distribution of the generated candidate masks. The target $Y$ is used for feature selection. Once the sampling mask has been determined, reconstruction takes the retained $k$-space data as input.

\subsection{Sampling-line evaluation using mutual and conditional mutual information}
MI characterizes statistical dependence, including nonlinear relationships, without requiring a prespecified functional form or parametric probability model. It is therefore widely used in feature selection.\nobreak\cite{battiti1994,peng2005}
For sampling line $i$, the MI between the selection variable \(X_i\) and the target variable $Y$ is estimated as\nobreak\cite{brown2012}
\begin{equation}
\widehat I(X_i;Y)
=\sum_{x_i,y}\widehat p(x_i,y)
\log\frac{\widehat p(x_i,y)}{\widehat p(x_i)\widehat p(y)},
\label{eq:new-mi}
\end{equation}
where \(\widehat p(x_i,y)\) is the empirical joint probability of \(X_i\) and $Y$, namely the fraction of candidate samples satisfying both \(X_i=x_i\) and \(Y=y\). The terms \(\widehat p(x_i)\) and \(\widehat p(y)\) are the corresponding empirical marginal probabilities. Here, \(x_i\in\{0,1\}\), and $y$ ranges over the observed values of the target variable.
Univariate MI measures the statistical association between each sampling line's selection state and the target variable $Y$. To account for conditional relationships between sampling lines, CMI evaluates how much of this association remains when the selection state of another line is given:
\begin{equation}
\widehat I(X_i;Y\mid X_j)
=\sum_{x_i,x_j,y}\widehat p(x_i,x_j,y)
\log\frac{\widehat p(x_i,y\mid x_j)}
{\widehat p(x_i\mid x_j)\widehat p(y\mid x_j)}.
\label{eq:new-cmi}
\end{equation}
Here, \(\widehat p(x_i,x_j,y)\) is the empirical joint probability of the three variables. The term \(\widehat p(x_i,y\mid x_j)\) is the conditional joint probability of \(X_i\) and $Y$ given \(X_j=x_j\), and \(\widehat p(x_i\mid x_j)\) and \(\widehat p(y\mid x_j)\) are the corresponding conditional marginal probabilities.
The CMI \(\widehat I(X_i;Y\mid X_j)\) describes the statistical association between the selection state of sampling line $i$ and the target variable, given the selection state of line $j$. Exchanging $i$ and $j$ gives the CMI in the other direction, \(\widehat I(X_j;Y\mid X_i)\). Both selected and unselected states of the conditioning variable are included and weighted by their empirical probabilities. All information measures are estimated from the empirical frequencies of the candidate samples using natural logarithms, with summation restricted to states with nonzero joint probability.
To evaluate pairs of sampling lines, we follow the feature-subset evaluation criterion of Pranji\'{c} et al.\nobreak\cite{pranjic2026} and account for the conditional associations in both directions through the pairwise score
\[
\widehat I(X_i;Y\mid X_j)+\widehat I(X_j;Y\mid X_i)
\]
for sampling lines $i$ and $j$. Each direction measures the statistical association conditioned on the selection state of the other line. Their sum gives a symmetric score independent of the ordering of $i$ and $j$. These scores measure association, not whether retaining a line reduces error; the resulting selection criterion is heuristic. The selection criterion thus combines the MI scores of individual sampling lines with pairwise CMI scores.

\subsection{QUBO formulation and solution}
\label{sec:qubo-solver}

The sampling mask is determined by optimizing the joint selection of sampling lines using these scores. As in Section~2.1, the binary variable \(z_i\) indicates whether sampling line $i$ is selected. Unlike the statistical variable \(X_i\) in the candidate dataset, \(z_i\) is a decision variable to be optimized. The MI and CMI estimates remain fixed during optimization.
Multiplying the MI score of each line by \(z_i\) includes that score only when the line is selected, while multiplying the pairwise CMI score by \(z_i z_j\) includes it only when both lines are selected. These contributions form the linear and quadratic terms, respectively. Maximization of the information scores is expressed as minimization of their negatives, and a squared penalty is added for deviations from the target number of sampling lines $k$:
\begin{equation}
\begin{aligned}
E(z)={}&-w_{\mathrm{MI}}\sum_{i=1}^{N}\widehat I(X_i;Y)\,z_i\\
&-w_{\mathrm{CMI}}\sum_{i<j}\Big[\widehat I(X_i;Y\mid X_j)
+\widehat I(X_j;Y\mid X_i)\Big]z_i z_j\\
&+\alpha\left(\sum_{i=1}^{N}z_i-k\right)^2,
\qquad z\in\{0,1\}^{N}.
\end{aligned}
\label{eq:new-qubo}
\end{equation}
The weights \(w_{\mathrm{MI}}\) and \(w_{\mathrm{CMI}}\) control the relative contributions of the univariate and pairwise scores. The condition \(i<j\) ensures that each pair is counted once, and \(\alpha>0\) controls the penalty for deviations from the target number of sampling lines. A squared-deviation penalty for a target selection cardinality is also used in the QUBO feature-selection model of Nau et al.\nobreak\cite{nau2025}.
With the variable transformation
\begin{equation}
z_i=\frac{s_i+1}{2},\qquad s_i\in\{-1,1\},
\label{eq:qubo-ising-map}
\end{equation}
the QUBO can be expressed as an Ising energy function, providing a model formulation suitable for annealing methods.\nobreak\cite{mucke2023}

We solve the same QUBO instances using SA and QA.
For SA, annealing samples are generated using multiple internal random seeds for each instance.
The lowest-energy configuration from each seed is then refined by local swaps: one selected line is exchanged with one unselected line, preserving the actual number of selected lines.
The squared penalty favors, but does not guarantee, exactly $k$ selected lines; swaps preserve the cardinality returned by each seed.
At each iteration, the swap producing the largest energy decrease is performed until the energy-decrease criterion is no longer satisfied.
Among the refined configurations from the internal seeds, the one with the lowest QUBO energy is selected as the representative sampling mask for subsequent reconstruction.
For QA, the returned masks are evaluated without additional local-swap refinement.
For both methods, test-set reconstruction metrics are not used to select the sampling masks.

\subsection{Image reconstruction and quality evaluation}
\label{sec:reconstruction-evaluation}

The sampling mask obtained by solving the QUBO is applied to the test set, and reconstruction is used to evaluate the feature-selection result. All sampling methods use the zero-filled reconstruction procedure in Eq.~\eqref{eq:candidate-zf}.
Let \(\widetilde u\) denote the reconstructed image and $u$ the reference image. Reconstruction quality is evaluated using RMSE\cite{wang2009mse}, peak signal-to-noise ratio (PSNR)\cite{wang2009mse}, and the structural similarity index (SSIM)\cite{wang2004}, which measure differences and similarities between the two images.
RMSE measures the pixelwise error and is defined as
\begin{equation}
\operatorname{RMSE}(u,\widetilde u)
=\sqrt{\frac{1}{NW}\sum_{a=1}^{N}\sum_{b=1}^{W}
\left(u_{ab}-\widetilde u_{ab}\right)^2}.
\label{eq:rmse-evaluation}
\end{equation}
The image and its Fourier grid both have dimensions $N\times W$. Each index along the first dimension of the Fourier grid identifies one sampling line, so $N$ is the total number of available lines in Eq.~\eqref{eq:target-sampling-budget}. A lower RMSE indicates a smaller pixelwise error. PSNR expresses this error on a logarithmic scale relative to a specified image signal range:
\begin{equation}
\operatorname{PSNR}(u,\widetilde u)
=20\log_{10}\frac{L}{\operatorname{RMSE}(u,\widetilde u)},
\label{eq:psnr-evaluation}
\end{equation}
where $L$ is the image signal dynamic range used for evaluation. PSNR is expressed in dB, with higher values indicating smaller pixelwise errors.
SSIM measures image similarity by comparing local luminance, contrast, and structure.\nobreak\cite{wang2004} The image-level SSIM is obtained by averaging the similarity over local windows:
\begin{equation}
\operatorname{SSIM}(u,\widetilde u)
=\frac{1}{|\Omega|}\sum_{q\in\Omega}
\frac{(2\mu_{u,q}\mu_{\widetilde u,q}+C_1)(2\sigma_{u\widetilde u,q}+C_2)}
{(\mu_{u,q}^{2}+\mu_{\widetilde u,q}^{2}+C_1)(\sigma_{u,q}^{2}+\sigma_{\widetilde u,q}^{2}+C_2)},
\label{eq:ssim-evaluation}
\end{equation}
Here, \(\Omega\) is the set of local windows used for evaluation. The quantities \(\mu_{u,q}\) and \(\mu_{\widetilde u,q}\) are the means within window $q$, \(\sigma_{u,q}^{2}\) and \(\sigma_{\widetilde u,q}^{2}\) are the corresponding variances, and \(\sigma_{u\widetilde u,q}\) is the covariance. The positive constants \(C_1\) and \(C_2\) stabilize the calculation. SSIM values closer to 1 indicate greater similarity between the reconstructed and reference images.
Supplementary experiments also apply total variation (TV) and wavelet-regularized reconstruction to the masks obtained from the QUBO. The mathematical models and solution methods are given in the Appendix~\ref{app:regularized-reconstruction}.

\subsection{Experimental settings}
\label{sec:experimental}

The experiments use a selected subset of the fastMRI single-coil knee dataset.\nobreak\cite{knoll2020} The training set contains 200 slices from 25 patients, and the test set contains 60 slices from 5 other patients. Reference images are obtained by taking the magnitude of the inverse Fourier transform of the selected slices' complex $k$-space data, center-cropping to $256\times256$, and normalizing each image. All sampling methods apply masks to the Fourier data of these reference images, selecting sampling lines along the first dimension of the Fourier grid without adding noise.

The target sampling rates are 15\%, 20\%, 25\%, 30\%, 35\%, and 40\%. At each rate, 1000 candidate masks are generated, comprising 500 uniform random candidates and 500 low-frequency Gaussian candidates. For a target sampling rate $r$, the low-frequency Gaussian candidates retain $n_{\mathrm{LF}}=\operatorname{round}(0.32rN)$ contiguous lines centered at zero frequency. The remaining lines are drawn from a Gaussian distribution centered at zero frequency with a standard deviation of 14 until the target number of sampling lines is reached. Uniform random candidates do not reserve a low-frequency region. Training reconstruction errors are discretized into 20 quantile intervals.

The QUBO information weights are $w_{\rm MI}=1$ and $w_{\rm CMI}=0.1$, and the sampling-count penalty weight is $\alpha=1$. SA is implemented using \texttt{dwave-neal}. For SA, each QUBO is solved with 5 internal random seeds, each using \texttt{num\_reads=500} and \texttt{num\_sweeps=500}. The representative mask is determined by the procedure in Section~\ref{sec:qubo-solver}.
\myrev{For QA, we use D-Wave's quantum--classical hybrid binary quadratic model (BQM) solver. The solver is accessed through \texttt{LeapHybridSampler}, with solver identifier \texttt{hybrid\_\allowbreak binary\_\allowbreak quadratic\_\allowbreak model\_\allowbreak version2p}.
The effective solver time limit is 3~s, the minimum allowed for these instances.
For each sampling rate and repetition, the hybrid solver uses the same QUBO coefficients as SA, without regenerating candidate masks or re-estimating MI and CMI.
The resulting masks are evaluated on the same test set using the same reconstruction procedures and metrics. The MI--CMI sampling methods using SA and QA are denoted QUBO-SA and QUBO-QA, respectively.}

The methods compared are QUBO-SA, QUBO-QA, univariate MI ranking (Top-$k$ MI), variable-density random sampling (VDS), Gaussian-weighted random sampling (Gaussian), uniform random sampling (Random), and approximately equispaced sampling (Equispaced).
Top-$k$ MI selects the $k$ lines with the highest MI scores.
VDS, Gaussian, Random, and Equispaced use the same contiguous low-frequency region as the low-frequency Gaussian candidates and differ only in how the remaining lines are selected.
VDS uses weighted sampling with
$w_i\propto[1+(d_i/(N/8))^2]^{-1}$,
where $d_i$ is the distance from sampling location $i$ to zero frequency.
Gaussian uses a broader Gaussian distribution ($\sigma=0.2N$) to cover a wider frequency range outside the retained low-frequency region.
Random samples uniformly at random outside that region, whereas Equispaced selects approximately equispaced lines outside it.
QUBO-SA, QUBO-QA, and Top-$k$ MI do not prescribe a contiguous low-frequency region.

Supplementary reconstruction experiments are conducted at a target sampling rate of 30\%. The QUBO-SA and QUBO-QA masks from each complete run are each held fixed while zero-filled, TV, and wavelet-regularized reconstruction are compared.
For all training-candidate and test-set evaluations, the reference and reconstructed magnitude images are separately min–max normalized, and the data range is set to 1 for both PSNR and SSIM.
The same fixed patient and slice split is used for all sampling rates and repeated experiments. At each sampling rate, the complete workflow of candidate generation, information estimation, optimization, and evaluation is repeated 5 times with seeds 42--46. Within each run, metrics are first averaged within patients and then aggregated with equal weights across test patients. Quantitative results are reported as the mean and sample standard deviation across complete runs. Differences between methods are paired within each run before aggregation. Quantitative evaluation uses the test set. Displayed images use fixed training slices, as indicated in the figure captions.

\section{Results}

\subsection{Reconstruction comparison with conventional sampling methods}

\begin{figure}[!htbp]
\centering
\includegraphics[width=\linewidth]{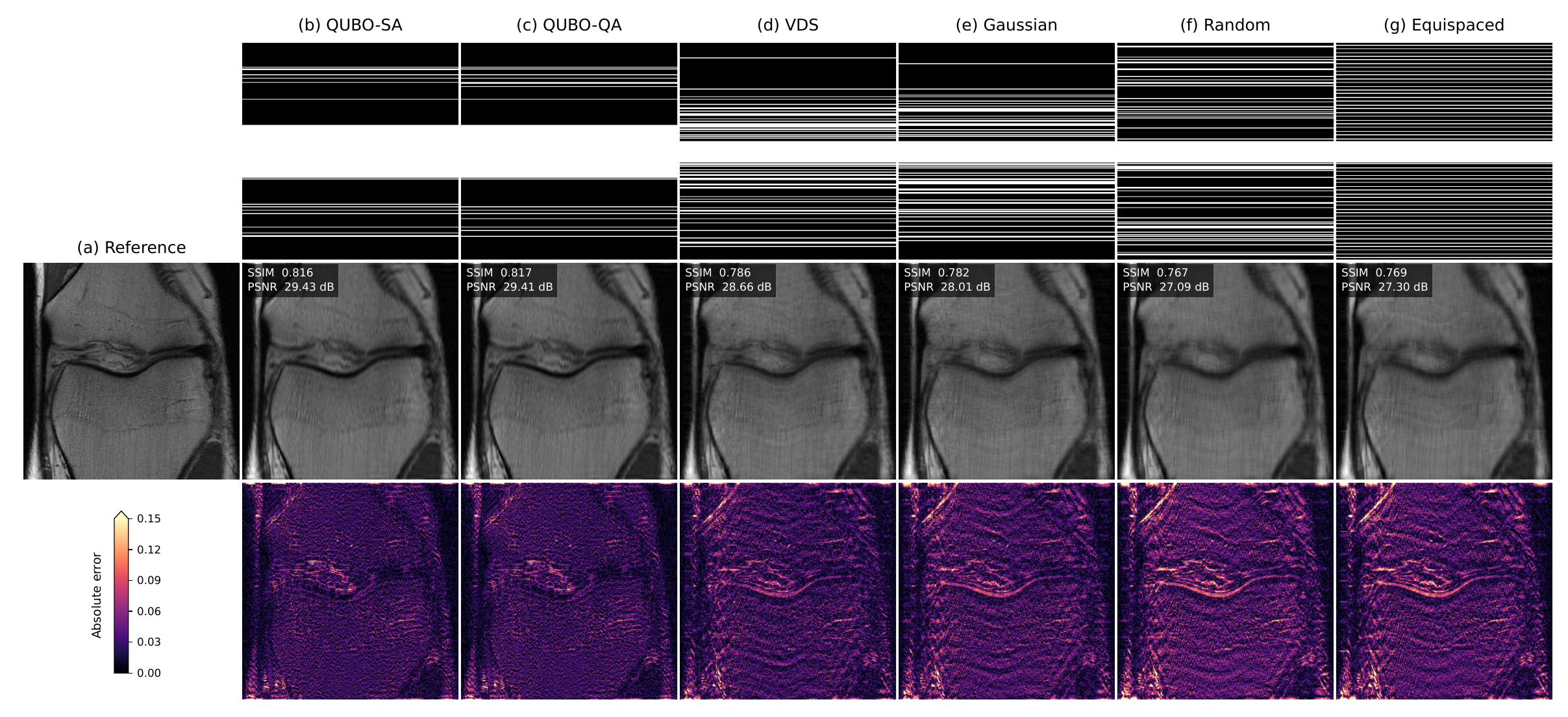}
\caption{Sampling masks and zero-filled reconstruction examples at a target sampling rate of 30\%. (a) Reference image; (b--g) QUBO-SA, QUBO-QA, VDS, Gaussian, Random, and Equispaced, respectively. Each column (b--g) shows, from top to bottom, the sampling mask, reconstructed image, and absolute error map. The example uses a prespecified training slice and the results of the third complete run; it is excluded from the independent test statistics, and the metrics shown in the figure apply only to this slice. Images use a common display range. The error color scale spans 0--0.15, with values above the upper limit displayed in the color corresponding to that limit.}
\label{fig:reconstruction-example}
\end{figure}

We first compared the reconstruction performance of QUBO-SA and QUBO-QA with four conventional MRI sampling methods VDS, Gaussian, Random, and Equispaced using the same zero-filled reconstruction procedure.
Figure~\ref{fig:reconstruction-example} presents a training-slice example at a target sampling rate of 30\%, showing the sampling mask, reconstructed image, and absolute error for each method. In this example, both QUBO-SA and QUBO-QA achieved higher SSIM and PSNR than all four conventional methods.

Table~\ref{tab:sampling30} summarizes reconstruction performance over the entire test set.
Both QUBO-SA and QUBO-QA achieved lower mean RMSE and higher mean SSIM and PSNR than all four conventional methods.
Their results were close, with QUBO-QA yielding the best mean values among these six methods for all three metrics.
The mean RMSE values were 0.05256 for QUBO-SA and 0.05252 for QUBO-QA, compared with 0.05879 for VDS, the best-performing conventional method by RMSE.
The relative reductions in mean RMSE compared with VDS were approximately 10.6\%--10.7\% for the two QUBO methods.

\begin{table}[!htbp]
\centering\small
\setlength{\tabcolsep}{4pt}\renewcommand{\arraystretch}{1.12}
\caption{Zero-filled reconstruction metrics for \myrev{QUBO-SA, QUBO-QA,} and four conventional sampling methods at a target sampling rate of 30\%. Values are means $\pm$ sample standard deviations across complete runs, with patients weighted equally within each run. The test set comprises 5 patients and 60 slices. Boldface denotes the best mean in each column. Equispaced uses a deterministic mask and therefore has zero standard deviation across runs.}
\label{tab:sampling30}
\small

\begin{tabular}{lrrr}
\toprule
Method & RMSE $\downarrow$ & SSIM $\uparrow$ & PSNR (dB) $\uparrow$ \\
\midrule
QUBO-SA & $0.05256\pm 0.00031$ & $0.7070\pm 0.0011$ & $26.19\pm 0.05$ \\
QUBO-QA & $\mathbf{0.05252}\pm 0.00030$ & $\mathbf{0.7073}\pm 0.0012$ & $\mathbf{26.20}\pm 0.04$ \\
VDS & $0.05879\pm 0.00130$ & $0.7035\pm 0.0071$ & $25.24\pm 0.20$ \\
Gaussian & $0.06050\pm 0.00100$ & $0.6990\pm 0.0046$ & $25.03\pm 0.15$ \\
Random & $0.06410\pm 0.00082$ & $0.6873\pm 0.0046$ & $24.58\pm 0.10$ \\
Equispaced & $0.06297\pm 0.00000$ & $0.6997\pm 0.0000$ & $24.70\pm 0.00$ \\
\bottomrule
\end{tabular}
\end{table}

\begin{figure}[!htbp]
\centering
\includegraphics[width=\linewidth]{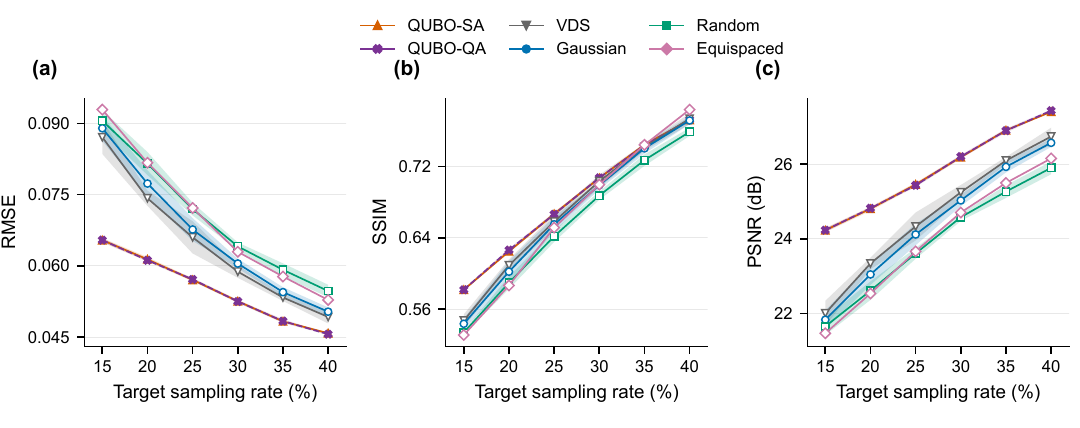}
\caption{Zero-filled reconstruction performance of QUBO-SA, QUBO-QA and four conventional sampling methods at different target sampling rates. (a) RMSE; (b) SSIM; (c) PSNR. Points and shaded bands show the mean and $\pm1$ sample standard deviation across complete runs, respectively, for metrics averaged with equal patient weights. The horizontal axis indicates the target sampling rate. Target rates are 15\%, 20\%, 25\%, 30\%, 35\%, and 40\%.}
\label{fig:sampling-rates}
\end{figure}

To examine how these differences vary with the sampling budget, we extended the comparison to target sampling rates of 15\%--40\% (Fig.~\ref{fig:sampling-rates}).
For all six methods, mean RMSE decreased and mean SSIM and PSNR increased as the sampling rate increased.

\myrev{QUBO-SA and QUBO-QA achieved lower mean RMSE and higher mean PSNR than all four conventional methods at all six target sampling rates.
The two QUBO methods showed similar trends across sampling rates, with closely matching mean values for all three metrics.
For both QUBO methods, the relative reduction in mean RMSE compared with VDS decreased as the target sampling rate increased.}
The ranking differed for SSIM. \myrev{At a target sampling rate of 40\%, Equispaced achieved the highest mean SSIM, exceeding both QUBO-SA and QUBO-QA.}
Thus, better performance in pixelwise error metrics did not always correspond to higher structural similarity.

\subsection{Comparison of MI ranking and the MI--CMI selection procedure}
\begin{figure}[!htbp]
\centering
\includegraphics[width=\linewidth]{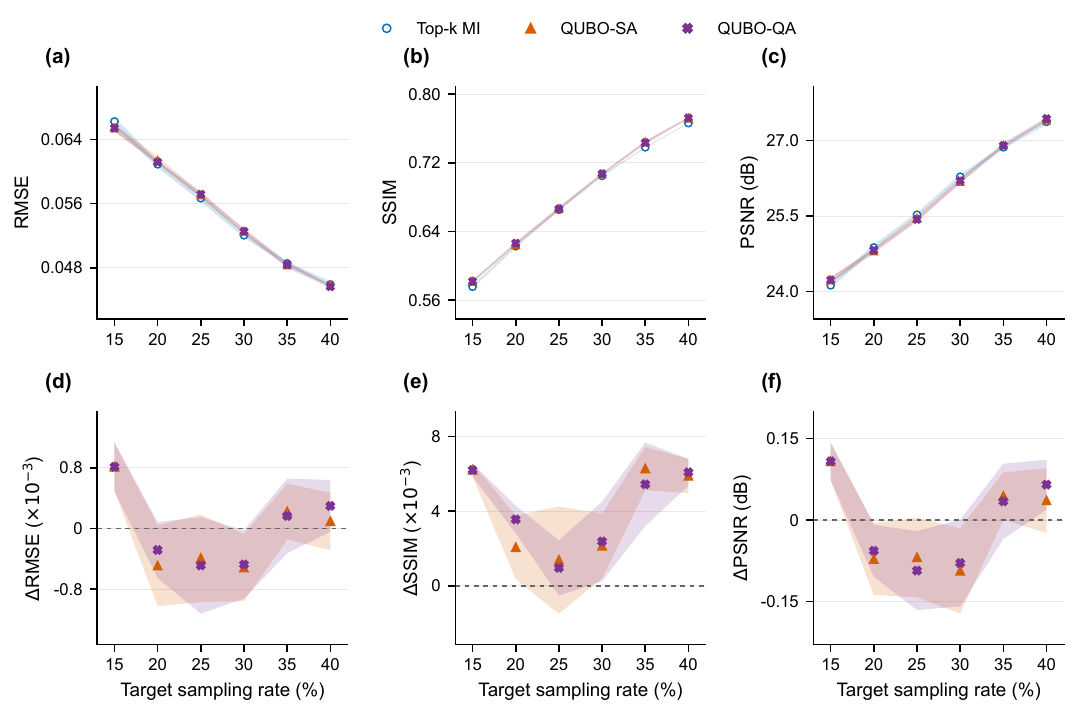}
\caption{Reconstruction metrics and paired differences for \myrev{Top-$k$ MI, QUBO-SA, and QUBO-QA.}
(a--c) RMSE, SSIM, and PSNR; points and shaded bands indicate the mean and $\pm1$ sample standard deviation across complete runs.
(d--f) Points and shaded bands indicate the mean and $\pm1$ sample standard deviation of within-run paired metric differences, respectively.
\myrev{For each QUBO method, $\Delta\mathrm{RMSE}$ is defined as Top-$k$ MI minus that method.
Conversely, $\Delta\mathrm{SSIM}$ and $\Delta\mathrm{PSNR}$ are defined as that method minus Top-$k$ MI.
Positive values therefore favor the corresponding QUBO method for all three metrics.}
Dashed lines indicate zero difference.}
\label{fig:mi-comparison}
\end{figure}

\myrev{To further examine reconstruction performance under joint selection relative to univariate ranking, we compared the MI--CMI selection procedure, implemented as QUBO-SA and QUBO-QA, with univariate Top-$k$ MI ranking (Fig.~\ref{fig:mi-comparison}).
At target sampling rates of 15\%, 35\%, and 40\%, QUBO-SA and QUBO-QA had lower mean RMSE and higher mean PSNR than Top-$k$ MI.
At 20\%--30\%, the ordering was reversed for these metrics.
Accordingly, the mean paired $\Delta\mathrm{RMSE}$ and $\Delta\mathrm{PSNR}$ values changed sign across sampling rates (Fig.~\ref{fig:mi-comparison}d,f).}
\myrev{For SSIM, the mean paired differences relative to Top-$k$ MI were positive at all six target sampling rates for QUBO-SA and QUBO-QA (Fig.~\ref{fig:mi-comparison}b,e).
These differences were larger at 15\%, 35\%, and 40\% than at 20\%--30\%.}

\subsection{Comparison of reconstruction methods with fixed sampling masks}

The preceding comparisons used the same zero-filled reconstruction procedure to evaluate reconstruction performance from the data retained by different sampling masks.
For a fixed sampling mask, regularized methods can also be used to reconstruct images from the same undersampled data.
We therefore fixed the \myrev{QUBO-SA and QUBO-QA masks separately} in each complete run at a target sampling rate of 30\% and compared zero-filled, TV, and wavelet-regularized reconstruction (Table~\ref{tab:reconstruction30}).
\myrev{For both QUBO-SA and QUBO-QA masks, TV reconstruction yielded lower mean RMSE and higher mean PSNR, but lower mean SSIM, than zero-filled reconstruction under the current parameter settings.
Wavelet reconstruction yielded higher mean RMSE and lower mean SSIM and PSNR than zero-filled reconstruction for both mask types.}

\begin{table}[!htbp]
\centering\small
\setlength{\tabcolsep}{4pt}\renewcommand{\arraystretch}{1.12}
\caption{\myrev{Comparison of reconstruction methods with fixed QUBO-SA and QUBO-QA masks at a target sampling rate of 30\%. Within each sampling method and complete run, zero-filled, total variation, and wavelet-regularized reconstruction were applied to the same mask and test images. Values are means $\pm$ sample standard deviations across complete runs, with patients weighted equally within each run. The test set comprises 5 patients and 60 slices.}}
\label{tab:reconstruction30}
\begin{tabular}{llrrr}
\toprule
Reconstruction method & Sampling method & RMSE $\downarrow$ & SSIM $\uparrow$ & PSNR (dB) $\uparrow$ \\
\midrule
\multirow{2}{*}{Zero-filled (ZF)} & QUBO-SA & $0.05256\pm 0.00031$ & $0.7070\pm 0.0011$ & $26.19\pm 0.05$ \\
 & QUBO-QA & $0.05252\pm 0.00030$ & $0.7073\pm 0.0012$ & $26.20\pm 0.04$ \\
\midrule
\multirow{2}{*}{Total variation (TV)} & QUBO-SA & $0.05052\pm 0.00047$ & $0.6914\pm 0.0018$ & $26.46\pm 0.06$ \\
 & QUBO-QA & $0.05050\pm 0.00044$ & $0.6915\pm 0.0024$ & $26.47\pm 0.03$ \\
\midrule
\multirow{2}{*}{Wavelet} & QUBO-SA & $0.05445\pm 0.00063$ & $0.6507\pm 0.0025$ & $25.80\pm 0.11$ \\
 & QUBO-QA & $0.05439\pm 0.00056$ & $0.6513\pm 0.0037$ & $25.81\pm 0.10$ \\
\bottomrule
\end{tabular}
\end{table}

\section{Conclusions and Discussion}

We formulate MRI undersampling mask design as feature-subset selection over Cartesian sampling lines. MI and bidirectional CMI define a criterion incorporating univariate and pairwise conditional associations with the target variable. 
These information scores are combined with a sampling-count penalty in a QUBO model, which is solved by SA and QA. 
The selected line subsets determine which $k$-space data are retained. 
Their corresponding masks are evaluated through image reconstruction on an independent test set of patients, linking information-theoretic feature selection to MRI sampling design.

\myrev{With zero-filled reconstruction, QUBO-SA and QUBO-QA had lower mean RMSE and higher mean PSNR than the four conventional sampling methods.
This ordering was observed at all six target sampling rates from 15\% to 40\%.
Relative to Top-$k$ MI, the ordering in mean RMSE and PSNR varied with sampling rate, whereas mean SSIM was higher for both QUBO methods at all six rates.}

The comparisons also showed that the relative performance of sampling methods depended on the evaluation metric.
\myrev{At a target sampling rate of 40\%, Equispaced had the highest mean SSIM.
QUBO-SA and QUBO-QA, however, had lower mean RMSE than Equispaced.}
Equispaced distributes peripheral sampling positions at approximately regular intervals outside the retained low-frequency region, which may help preserve local image structure.
A similar divergence between metrics was observed when reconstruction methods were compared using fixed \myrev{QUBO-SA and QUBO-QA} masks at a target sampling rate of 30\%.
\myrev{For both sampling methods under the current parameter settings, TV yielded lower mean RMSE and higher mean PSNR, but lower mean SSIM, than zero-filled reconstruction.
Wavelet reconstruction yielded higher mean RMSE and lower mean SSIM and PSNR than zero-filled reconstruction for both sampling methods.
Lower pixelwise error was therefore not always accompanied by higher structural similarity in these sampling and reconstruction comparisons.}

The present evaluation used data from a limited number of knee MRI patients and simulated undersampling using Fourier data generated from preprocessed magnitude images.
Because each reference and reconstruction is normalized separately before evaluation, the reported metrics do not assess absolute signal-scale fidelity.
Future studies could extend the evaluation to more patients, different acquisition conditions, and raw complex-valued $k$-space data to assess the generalizability of these findings.
Building on the comparisons with conventional sampling methods and univariate MI ranking, future work could also include learning-based sampling mask optimization methods.
Evaluations under comparable conditions could clarify their relative reconstruction performance, computational cost, and applicability.
\myrev{The present study used SA and a quantum--classical hybrid solver to obtain sampling masks from the QUBO formulation.
The evaluation focused on the reconstruction performance of the resulting masks.
A systematic comparison of solver computational cost and QUBO solution quality remains for future work.
Together, the formulation and reconstruction experiments indicate that information-theoretic feature selection can provide an effective approach to MRI sampling-mask design under the tested simulation settings.}

\begin{acknowledgment}
We received financial support from the Cross-ministerial Strategic Innovation Promotion Program (SIP) of the Cabinet Office
(No. 23836436).
This work was also supported by JST BOOST, Japan Grant Number JPMJBS2421.
\end{acknowledgment}

\appendix
\section{Regularized reconstruction models and solvers}
\label{app:regularized-reconstruction}
\setcounter{equation}{0}
\renewcommand{\theequation}{\Alph{section}.\arabic{equation}}

Regularized reconstruction uses the centered, unitarily normalized two-dimensional discrete Fourier operator $\mathcal F_{\rm u}$. For a reference magnitude image $u$ and a fixed sampling-line selection vector $z$, we define $A^{\rm u}_zv=\mathcal M(z)\odot\mathcal F_{\rm u}v$ and generate the data using the same normalization convention, $b^{\rm u}_z=A^{\rm u}_zu$. Zero-filled reconstruction is given by
\begin{equation}
\widehat u_{\rm ZF}=\mathcal F_{\rm u}^{-1}b^{\rm u}_z.
\label{eq:new-zf}
\end{equation}
The zero-filled implementations used for training-candidate evaluation and the main sampling comparison employ a conventional paired forward and inverse Fourier transform. Their normalization factors cancel in the composite operation, yielding the same complex-valued zero-filled image as Eq.~\eqref{eq:new-zf}, before magnitude extraction and image normalization.. Regularized reconstruction consistently uses the unitary normalization convention above to specify the relative scales of the data-fidelity and regularization terms.

Isotropic total variation (TV) and wavelet reconstruction are defined, respectively, as
\begin{equation}
\widehat u_{\rm TV}\in\arg\min_{v\in\mathbb C^{N\times W}}
\frac12\|A^{\rm u}_zv-b^{\rm u}_z\|_2^2
+\lambda_{\rm TV}\sum_p\sqrt{|D_1v_p|^2+|D_2v_p|^2},
\label{eq:new-tv}
\end{equation}
\begin{equation}
\widehat u_{\rm W}\in\arg\min_{v\in\mathbb C^{N\times W}}
\frac12\|A^{\rm u}_zv-b^{\rm u}_z\|_2^2+\lambda_{\rm W}\|P_d\Psi v\|_1.
\label{eq:new-wavelet}
\end{equation}
Here, $D_1,D_2$ denote forward differences in the two spatial directions, $p$ indexes pixels, $\Psi$ is an orthogonal wavelet transform, and $P_d$ extracts the detail coefficients. The two regularizers exploit sparsity in image gradients and wavelet coefficients, respectively.\nobreak\cite{lustig2007}

TV reconstruction is solved using the alternating direction method of multipliers (ADMM),\nobreak\cite{boyd2011} with $\lambda_{\rm TV}=10^{-3}$, an ADMM penalty parameter $\rho=0.1$, and periodic boundary conditions for the difference operators. Iteration stops when both the primal and dual residuals satisfy their stopping thresholds, with an absolute tolerance of $10^{-5}$ and a relative tolerance of $10^{-4}$. Wavelet reconstruction uses the fast iterative shrinkage-thresholding algorithm (FISTA),\nobreak\cite{beck2009} with $\lambda_{\rm W}=10^{-3}$, a three-level orthogonal db4 wavelet transform with periodized boundary handling, a penalty applied only to the detail coefficients, and a step size of 1. Iteration stops when the proximal gradient mapping residual, divided by $\max(1,\|v\|_2)$, is no greater than $10^{-5}$. Both methods are initialized with the zero-filled reconstruction and run for a maximum of 500 iterations, with identical numerical settings across all complete runs. Reconstruction magnitudes are normalized and evaluated according to the procedure in Section~\ref{sec:experimental}.

\bibliographystyle{unsrt}
\bibliography{references}

\end{document}